\documentclass{article}

\usepackage[numbers]{natbib}
\usepackage[final]{neurips_2019}

\usepackage[utf8]{inputenc} 
\usepackage[T1]{fontenc}    
\usepackage{hyperref}       
\usepackage{url}            
\usepackage{booktabs}       
\usepackage{amsfonts}       
\usepackage{nicefrac}       
\usepackage{microtype}      
\usepackage{graphicx}
\usepackage{subcaption}

\title{Unsupervised deep clustering for predictive texture pattern discovery in medical images}

\author{%
    Matthias Perkonigg
  \And
  Daniel Sobotka
  \And
  Ahmed Ba-Ssalamah
  \And
    Georg Langs \vspace{4mm}\\
  Computational Imaging Research Lab (CIR) \\
  Department of Biomedical Imaging and Image-guided Therapy \\
  Medical University of Vienna, Austria\\
}

\begin{document}

\maketitle

\begin{abstract}
  Predictive marker patterns in imaging data are a means to quantify disease and progression, but their identification is challenging, if the underlying biology is poorly understood. Here, we present a method to identify predictive texture patterns in medical images in an unsupervised way. Based on deep clustering networks, we simultaneously encode and cluster medical image patches in a low-dimensional latent space. The resulting clusters serve as features for disease staging, linking them to the underlying disease. We evaluate the method on 70 T1-weighted magnetic resonance images of patients with different stages of liver steatosis. The deep clustering approach is able to find predictive clusters with a stable ranking, differentiating between low and high steatosis with an F1-Score of 0.78.
\end{abstract}

\section{Introduction}

Marker patterns serve prediction and quantification of disease status, and their machine learning based discovery is relevant if the underlying biological processes are poorly understood, or no a priori defined markers are known. 
Here, we pose the identification of marker pattern candidates in imaging data of patients with a particular disease as a simultaneous dimensionality reduction and clustering problem. We perform unsupervised deep clustering of texture patterns in medical images and identify those patterns linked to the stages of an underlying disease present in the population. The discovery of frequent image patterns contributing to classification enables the analysis of regions where those patterns occur, and is more interpretable than global image features.
Several relevant approaches exist~\cite{caron2018,yang2017,yang2016jule,DeBrabandere}, and unsupervised clustering of deep features in the medical domain has shown promise for unsupervised segmentation \cite{moriya2018unsupervised} or image categorization \cite{wang2017unsupervised}. We build on Deep Clustering Networks (DCN) jointly learning and clustering latent space features instead of performing these steps sequentially~\cite{yang2017}. 

We perform deep convolutional clustering on medical image patches (image segments) in a real-world problem setting to find distinct, predictive texture patterns that can be used for the staging of liver steatosis in MRI data. Steatosis is an accumulation of fat within the liver. It is one of the characteristics of Nonalcoholic Fatty Liver Disease (NAFLD) \cite{Kleiner2005}. Steatosis is graded into four stages (Grade 0: \textless 5\%, Grade 1: 5\%-33\%, Grade 2: 33\%-66\%, Grade 3: \textgreater 66\%) and can be assessed by biopsy, ultrasound or MRI-Proton Density Fat Fraction (PDFF) and MR elastography (MRE) \cite{Dulai2016}. In this work we link texture patterns in T1 weighted magnetic resonance imaging (MRI) data discovered by our unsupervised approach to the steatosis grade. 

\section{Methods}
The proposed method can be divided into four major steps: First, we randomly extract 2D axial patches within an anatomical structure - in this case the liver - from image volumes. Then, we simultaneously encode and cluster appearance variability in these patches in a latent space with DCN, and apply the trained DCN to calculate a signature of relative cluster proportions for each case. Finally, we link signatures to disease staging by training a random forest classifier~\cite{breiman2001random}, and scoring the contribution of each cluster to classification accuracy.

\paragraph{Encoding and clustering of image patches with convolutional DCNs}

Building upon the DCN~\cite{yang2017} approach we use a deep convolutional autoencoder to encode a set of image patches in a latent space. Three convolutional layers (50, 20 and 10 feature maps respectively) with receptive field of 3x3 and ReLU activation followed by a 2x2 MaxPooling are used as an encoder to a 20 dimensional latent space. Deconvolution in the decoder is performed by upsampling followed by a convolutional layer in the reversed order. In addition to the reconstruction loss of the conventional autoencoder, DCN uses a second loss term  measures the 'cluster-friendliness' derived from a k-means algorithm \cite{lloyd1982least}. The full loss function is
\begin{equation}
\min_{\mathit{W}, \mathit{Z}, \mathit{M}, \{s_i\}} \sum_{n=1}^N (l(\mathbf{g}(\mathbf{f}(x_i), x_i) + \lambda \Vert \mathbf{f}(x_i) - \mathit{M}s_i\Vert_2^2,
\end{equation}
where $\mathit{W}$ and $\mathit{Z}$ are the parameters of the encoder ($\mathbf{f}$) and decoder ($\mathbf{g}$) respectively. $\mathit{M}$ are the centroids of the clusters and $\{s_i\}$ is the cluster membership of patch $i$. $l$ is a reconstruction loss function, in our approach we use the mean-squared error. $\lambda$ is a hyperparameter that regulates the trade-off between reconstruction quality and 'cluster-friendliness'~\cite{yang2017}. 

The network is trained in an alternating gradient descent scheme of (1) updating the network parameters, (2) updating the cluster assignments of the training patches and (3) updating the cluster centroids~\cite{yang2017}.

\paragraph{Calculation of the signature of an image region} 

After training the convolutional DCN (cDCN), we apply it to a whole region of interest (ROI, e.g., anatomical structure) by parsing it with a sliding window, and assigning each position to one of the learned clusters. For each ROI the vector of the relative proportion of each cluster forms the \textit{signature} of the ROI. This approach is similar to Bag-of-Visual-Words~\cite{sivic2003video}.

\paragraph{Linking signatures to clinical parameters} 

If signature components are linked to clinical parameters, they hold meaningful pattern variations of the ROI. We calculate the signatures for all ROIs in the training data, and train a classifier to predict steatosis grade from the signature. We used Random Forests~\cite{breiman2001random} and LASSO regression~\cite{tibshirani1996regression} to quantify the link of individual signature components to disease stage, by quantifying its contribution to classifier accuracy, yielding an interpretable set of predictive components. 

\section{Results}

The method is evaluated on multi-center data of T1-weighted MRI volumes of 70 NASH patients for whom diagnosis and true steatosis grades are confirmed by biopsy (grade 0/1/2/3; n=1/20/32/17). 
We evaluate classification of individual grades and of low/high steatosis (grades 0\&1/2\&3). 
It is worth noting that in T1-weighted MRIs, other than in e.g. MRI-PDFF, the steatosis grade is not directly quantifyable~\cite{Dulai2016}. By linking signatures of T1-weighted MRIs to steatosis grade, we demonstrate that our approach can identify patterns that capture disease progression, even if no a priori known marker exists in the data.
Prior to analysis, livers are segmented with a U-Net \cite{ronneberger2015u} and masks are manually corrected. For training the cDCN 50.000 image patches with a size of 14x14mm (resampled to 32x32 pixels) are extracted at random positions within the livers of the dataset. We train the DCN to cluster the patches into 10 clusters. The number of clusters is a hyperparameter of DCN and is fixed after preliminary experiments.
\begin{figure}[!t]
  \centering
  \begin{minipage}[b]{0.42\linewidth}
    \includegraphics[width=\linewidth]{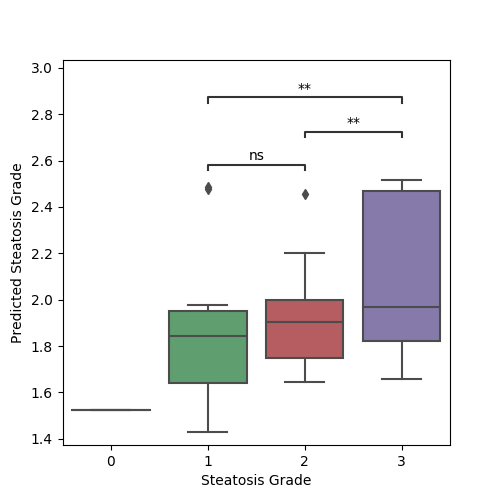}
    \caption{Regression values with corresponding ground truth steatosis grades.}
    \label{fig:regressionvalues}
  \end{minipage}
  \hspace{0.15cm}
  \begin{minipage}[b]{0.54\linewidth}
    \includegraphics[width=\linewidth]{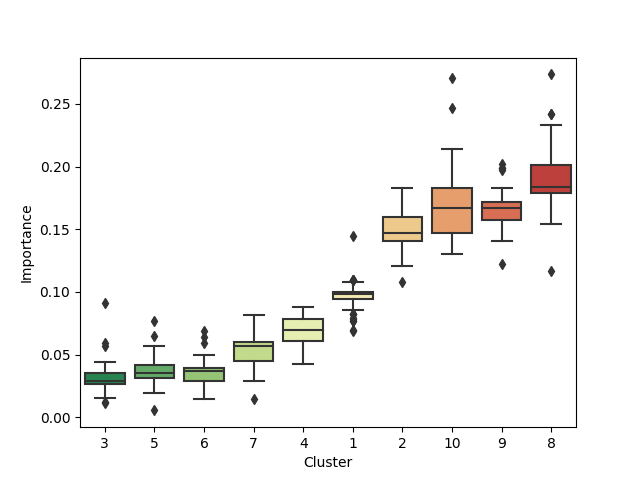}
    \caption{Feature Importance of the random forest classifier for low/high steatosis classification.}
    \label{fig:featureimportance}
  \end{minipage}
\end{figure}
Signatures for all MRI volumes are calculated within the liver. We applied classification and predictive signature component identification using LASSO regression and a random forest classifier using a Leave-One-Out cross validation scheme. 

\begin{figure}[b]
  \centering
    \includegraphics[width=\textwidth]{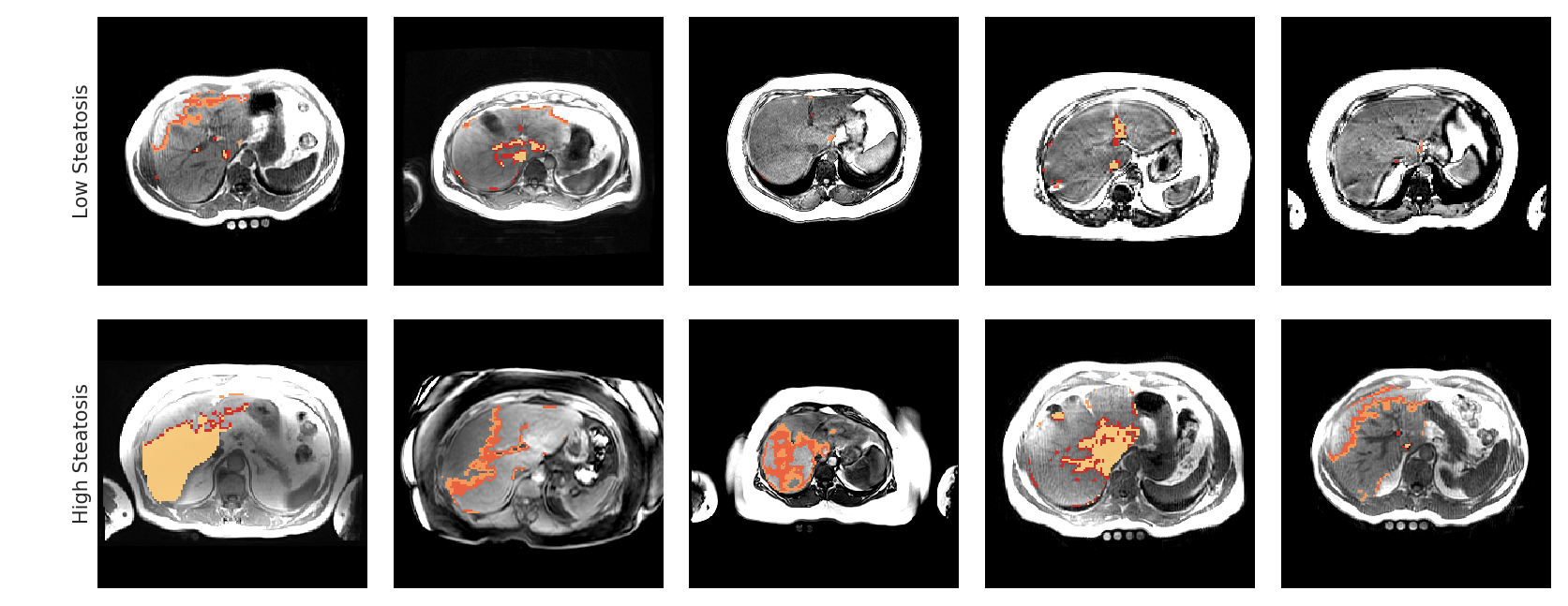}
  \caption{Examples of a low and a high steatosis patients with overlays showing the distribution of the four most informative signature components in these cases.}
  \label{fig:importantsamples}
\end{figure}

The results of predicting grade from signature are shown in Figure \ref{fig:regressionvalues}. Steatosis grade 3 can be reliably separated from all other grades and predicted values are correlated with true steatosis grade estimated from biopsy. Low/high steatosis classification reaches an accuracy of 0.70, Sensitivity of 0.78, Specificity of 0.52 and an F-Score of 0.78. Figure \ref{fig:featureimportance} shows the feature importance and its variability across leave-one-out samples in the population for a random forest classifier between low/high steatosis. Clusters 2, 10, 9 and 8 are consistently the most important features in the Leave-One-Out cross validation. To illustrate the presence and distribution of signature components, in Figure \ref{fig:importantsamples} those four clusters are shown on MR images from patients with different grades of steatosis. Two of the main predictive components (8 and 9) focus on regions around vessels. 

\section{Discussion}

We applied unsupervised clustering to a real-world problem of finding predictive features in medical imaging. We show how clusters can be linked to clinical parameters and demonstrate that the identified predictive signature components are stable and interpretable by humans.

We link the signature components identified in T1w MRI data to steatosis grade observed in biopsy. This grade can be non-invasively assessed \cite{Dulai2016}, but our results demonstrate that cDCN can identify predictive features in data where no a priori markers are known. Ground-truth steatosis grade based on biopsy is limited to a single position in the liver, obtained at a possibly different time point, while we assess the steatosis grade over the whole liver, possibly leading to discrepancy due to heterogeneity in the liver. 
Future work will extend our approach to multi-parametric MRI and link clusters to other clinical parameters such as inflammation. Aside from linking signature components to known changes associated with a disease, they offer the possibility to identify new hypotheses regarding imaging biomarker candidates. 

\subsubsection*{Acknowledgments}

This work was supported by the Austrian Science Fund (FWF) I 2714B31, OENB 18207 and Novartis Pharmaceuticals Corporation.  

\small
\bibliographystyle{plain}
\bibliography{unsupervised} 
\end{document}